\documentclass[submission,copyright]{eptcs}
\providecommand{\event}{FMAS 2021} 
\usepackage{underscore}  
\usepackage{amsmath}
\usepackage{subfigure}
\usepackage{amssymb}
\usepackage{capt-of}
\usepackage{listings}
\usepackage{array}
\usepackage[utf8]{inputenc}
\usepackage{float}
\usepackage[title]{appendix}
\usepackage{verbatim}
\usepackage{graphicx}
\usepackage{color}
\usepackage[dvipsnames]{xcolor}
\title{Simulation and Model Checking for Close to Real-time Overtaking Planning}
\author{Daumantas Pagojus
\institute{School of Computing Science\\
University of Glasgow}
\email{daumantas.pagojus@gmail.com}
\and
Alice Miller 
\institute{School of Computing Science\\
University of Glasgow}
\email{alice.miller@glasgow.ac.uk}
\and 
Bernd Porr 
\institute{School of Biomedical  Engineering\\
University of Glasgow}
\email{bernd.porr@glasgow.ac.uk}
\and 
Ivaylo Valkov 
\institute{School of Computing Science\\
University of Glasgow}
\email{ivaylo.valkov@glasgow.ac.uk}
}
\def\titlerunning{Simulation and Model Checking for Close to Real-time Overtaking Planning}
\def\authorrunning{D. Pagojus, A. Miller, B. Porr and I. Valkov}
\begin{document}
\lstdefinestyle{myCustomMatlabStyle}{
  language=Matlab,
  numbers=left,
  stepnumber=1,
  numbersep=10pt,
  tabsize=4,
  showspaces=false,
  showstringspaces=false
}
\maketitle

\begin{abstract}
 Fast and reliable trajectory planning is a key requirement of autonomous vehicles. In this paper we introduce a novel technique for planning the route of an autonomous vehicle on a straight rural road using the Spin model checker. We show how we can combine Spin's ability to identify paths violating temporal properties with sensor information from a 3D Unity simulation of an autonomous vehicle, to plan and perform consecutive overtaking manoeuvres on a traffic-heavy road. This involves discretising the sensory information and combining multiple sequential Spin models with a Linear Time Temporal Logic specification to generate an error path. This path provides the autonomous vehicle with an action plan. The entire process takes place in close to real-time (using no pre-computed data) and the action plan is specifically tailored for individual scenarios. Our experiments demonstrate that the simulated autonomous vehicle implementing our approach can drive on average at least 40 km and overtake 214 vehicles before experiencing a collision - which is usually caused by inaccuracies in the sensory system. While the proposed system has some drawbacks, we believe that our novel approach demonstrates a potentially powerful future tool for efficient trajectory planning for autonomous vehicles.

\end{abstract}

\section{Introduction}
\subsection{Motivation}
	Autonomous vehicles have been shown to be useful in a variety of tasks in both military and commercial contexts. The U.S. Office of Naval Research has demonstrated how a swarm of unmanned boats can help to patrol harbours \cite{hsu2019}, NASA has landed two autonomous rovers on Mars \cite{ackerman2013, ackerman2021} to search for proof of life on the planet, and Waymo are expanding their autonomous taxi service \cite{harris2018}.
	
	These systems need to be able to {\it see} the surrounding environment in a precise way, reason about the risks associated with each possible action and plan a collision free path for the vehicle to take. Accidents like the fatal Tesla crash using an autopilot mode \cite{yadron2016} or the fatal Uber autonomous taxi crash \cite{conger2020} show that current state-of-the-art systems are not yet capable of doing this task without error.

    In this paper, we investigate the decision making process of an autonomous vehicle.  Specifically, we use the Spin model checker \cite{holz2011} for overtaking planning.  We explore a scenario in which an autonomous vehicle equipped with sensors evaluates the environment in real-time and uses model checking to make hard decisions on what actions it should take to successfully complete multiple overtaking manoeuvres. We use simple models of relationships between the autonomous vehicle's actions and other traffic participants' behaviour, rather than relying on complex, computationally intensive algorithms or preprocessed data.
    
    \subsection{Approach}
    Our system uses a model of a vehicle, described using the Promela model specification language, which is analysed using the Spin model checker \cite{holz2011}. This model allows one to describe the surrounding traffic participants' behaviour for each action the vehicle can take in an overtaking scenario. Spin allows us to analyse this behaviour and provide the vehicle with a collision-free path to take to overtake another vehicle.
	We use a Unity simulator to test this model. Our simulation scenario mimics a straight rural two-way road with traffic in both lanes (on their left-hand side in each case - see Figure \ref{finalmovementexample}). The aim is for the autonomous vehicle to use information supplied by our model checking algorithm to overtake another vehicle while avoiding collisions with other vehicles. Realistic LiDAR and laser based sensors are simulated for the autonomous vehicle to provide environmental information to be used by the model checking program. 
	Model checking \cite{baier} is a widely used technique for automatically verifying reactive systems. It is based on a precise mathematical and unambiguous model of the possible system behaviour. To verify that the model meets specified requirements (usually expressed in temporal logic), all possible executions are checked systematically. If an execution failing the specification is found, the execution path which caused the violation is returned. 
	Model checking has previously been successfully used in a variety of different systems. It helped to ensure the safety and reliability of safety-critical systems like flight control \cite{Wang2018},  space-craft controllers \cite{Havelund2001-3} and satellite positioning systems \cite{lu2015}. Model checking has traditionally been used to detect software bugs and for protocol analysis. However, it has also been used to solve problems like detecting computer worms \cite{Kinder2010} and balancing turn-based games \cite{Kavanagh2019}. 

\subsection{Related Work}
	Currently, autonomous driving algorithms can be divided into two main categories: offline and online. Offline algorithms use pre-computed data to predict future actions of surrounding vehicles and decide which actions to take to achieve the specified goal. As an example, convolutional neural networks based algorithms \cite{deo2018} use images from highway cameras to train a model which can then predict the movements of all vehicles on the road and generate actions for the autonomous vehicle to execute. One problem with this approach is that the footage used to train the model is not necessarily recorded from the autonomous vehicle itself. This could cause problems once real data is used as input to the model when driving. Another example is a Deep Reinforcement Learning based algorithm \cite{Josef2020} which has been shown to be viable for planning a trajectory for a ground vehicle in an unknown terrain through the use of a sophisticated reward system. 
	
	The advantages of such an approach are that the data analysis takes place before driving occurs. While driving, path planning is very fast, as it is then simply a classification task. However, it is impossible to include all traffic situations in training data due to the high unpredictability of the environment (other road users, weather and road conditions for example). This could lead to incorrect prediction and, as a result, unexpected or risky actions taken by the autonomous vehicle. 
	
	In online algorithms, all computation takes place in real-time, for all road scenarios. As such, the resulting vehicle actions are tailored for the specific situation, and are thus oftern more accurate and safe. The downside however is the execution time. Since vehicles move at high speeds, a fast algorithm is required so that predictions are not immediately out of date.

	One of the most popular approaches for planning an autonomous vehicle's trajectory is Model Predictive Control (MPC). This approach has been used to model human-driver like control in various traffic simulations \cite{Ji2017, Prokop2001}. It has also been used \cite{Wang2019} to solve the problem of an autonomous vehicle getting stuck in a lane by forcing it to change into an adjacent empty lane or into a lane with assigned speed similar to the cruise control speed of the autonomous vehicle. In \cite{Rasekhipour2017} the representation of MPC obstacles were altered to include a wider range of dynamic features. 
	
	There have been successful experiments performed using a hierarchical framework proposed in \cite{li2016} which uses a two stage real-time trajectory planner. In the first stage the high-level behaviour planner makes a decision on what action to do - follow lane, change lane, overtake, etc. In the second stage a collision-free trajectory is generated for executing the provided action. Our proposed system is based on a similar idea - we generate actions (e.g. change lane) which we then pass to the autonomous vehicle to execute. 
	
	Online algorithms require prediction of other vehicles' behaviour to avoid a collision. To do this in real-time, methods like Markov Chains and Monte Carlo Simulations \cite{Althoff2011} or reachability analysis \cite{Althoff2014} can be used.

    There have been numerous formal approaches presented that are relevant for the analysis of self-driving cars. For example, in \cite{damopera2018}, a visual specification language: Traffic Sequence Charts, is proposed for the formal description of traffic-based scenarios within an autonomous context. 
     In \cite{kademcfive2017} vehicle platooning is represented as a multi-agent system using the GWENDOLEN agent programming language. Agent behaviour and real-time requirements of the system are verified using Agent Java PathFinder (AJPF) \cite{defiwebor2012} and the UPPAAL model checker \cite{lapeyi97}  respectively. A similar approach is used in 
    \cite{fecualfi2017} to verify the behaviour and plans of a rational agent acting as decision maker for an autonomous vehicle. 
    
    Model checking and simulation is used  for trajectory planning  in \cite{frgihoirmino2020}. The approach described there is for unmanned aerial vehicles rather than self-driving cars, and uses probabilistic models for route planning. The goals and the context in which our vehicle operates is entirely different. We are concerned with using live model checking in a simulated dynamic environment for safe overtaking of ground vehicles.  Recent work \cite{alnuwiquaive2021} uses model checking combined with sensor data for the verification of decision making in self-driving cars, as we do. Their focus is on autonomous parking rather than overtaking, and they use two different model checkers (MCMAS \cite{loqura2009} and PRISM \cite{kwiatkowska}), whereas we use Spin.
    Overtaking manoeuvres in the presence of oncoming traffic are modelled in \cite{hiliol2013}, using Multi-lane Spatial Logic (MLSL). 
    
    \subsection{Contributions}
    Our contributions are:
    \begin{itemize}
        \item A Promela model describing the behaviour of an autonomous vehicle and other road users.
        \item We describe how continuous sensory data from a (simulated) autonomous vehicle can be discretised and combined with our Promela model.
        \item An explain of how the Spin model checker is used to analyse our models and provide the autonomous vehicle with a list of instructions for overtaking other vehicles. 
        \item Results of testing our proposed system in a Unity simulation and suggestions future improvements.
    \end{itemize}
    
	The paper is structured as follows. In Section \ref{background} we introduce the key technologies used in our system, namely the Spin model checker, Linear Time Temporal Logic (LTL) and Unity. In Section \ref{methodology} we describe our overall methodology and in Sections \ref{sect:promelamodels} and \ref{sect:unitymodels} we present our Promela models and Unity Simulator. In Section \ref{usingmodelchecking} we provide details of testing performed, results achieved and our observations for improvements to our system. We conclude in Section \ref{conclusion}.

\section{Background}
\label{background}
 Our system has two major components: a Promela model to describe the autonomous vehicle and the surrounding traffic, which is analysed using the Spin model checker; and a simulator implemented in Unity from which we detect sensory data and to which we provide instructions for overtaking.  We introduce these two technologies in this section. 

\subsection{Promela and Spin}\label{sect:promela}
Model checking \cite{clagrpe99} is a method for verifying whether a finite-state model of a system meets a given logical specification (property). An abstract system description (model) is first specified using a model specification language, and the underlying model (transitions system), is analysed using a purpose built engine, such as Spin \cite{holz2011}, PRISM \cite{kwiatkowska} or Uppaal \cite{lapeyi97}. The analysis involves exploring the states and paths of the underlying transition system to determine whether the logical property holds.

In this paper we use the Spin model checker \cite{holz2011}. Spin is a widely used open-source software verification tool. It can be used for the formal verification of multi-threaded software applications and supports the high-level state-based description language Promela  
(PROcess MEta LAnguage) which is loosely based on Dijkstra's guarded command language \cite{dijkstra76}. The core aspects of Promela are:

\begin{itemize}
\item Promela specifications consist of process templates called {\em proctypes}, global message channels and other global variables. 
\item Each component type is declared within a \texttt{proctype} structure which contains local variables and statements. 
\item Each proctype instantiation represents the behaviour of an individual component. 
\item Guards (statements which may block, like boolean statements or channel operations) and choice statements control execution flow.  
\item Inter-component communication takes place via shared variables and channels. Apart from initialisation, our Promela models represent only a single process (the autonomous vehicle - all other vehicles are reflected by their effect on $AV$, modelled within this process), so we do not describe channel communication here. 
\item Two choice constructs (\texttt{if\ldots fi} and \texttt{do\ldots od} statements) allow  us to express non-determinism in our model.  
\item Variable types include \texttt{bit}, \texttt{byte}, \texttt{short}, \texttt{int}, \texttt{array} and an enumerated type \texttt{mtype}.
\item Promela programs contain initialisation information - i.e. how many and which particular instantiations of each proctype to create when the program is run.  
\end{itemize}

When a Promela specification is compiled and run with Spin, a global automaton consisting of all of the initiated components is constructed and combined with an automaton capturing any included LTL property. This merged automaton is referred to as the underlying {\t state-space} of the model. By searching the state-space, Spin can capture any executions of the model that violate the property.  As a model's complexity increases (e.g. with the number of components or number of local states per component) the size of state-space grows accordingly. Indeed, the size of the state-space grows exponentially with the number of components.  This phenomenon is known as the ``state-space explosion problem" \cite{Clarke2001} and means that the art of model checking is to keep models as abstract as possible, whilst maintaining their integrity as a representation of a physical system. 

Spin supports the specification of properties in Linear Time Temporal Logic (LTL) \cite{pnueli81}. LTL properties consist of a finite set of atomic propositions, boolean connectives, and  temporal operators, e.g.  \lstinline{<>} (``eventually") and  \lstinline|[]| (``always"). We do not provide details here, except to describe the single property that we use in this paper. This property is \lstinline{!<>p} which states that, from the initial state, in all paths, $p$ is never true (note that "for all paths" is implicit for all LTL properties). We will use this property to capture paths where $p$ does become true (where $p$ is defined to be a proposition which indicates that a successful overtake procedure has taken place).

\subsection{Unity}
Unity is a platform for creating and operating interactive, real-time 3D content\footnote{\url{https://unity.com/products/unity-platform}}. We have chosen Unity because of its in-built physics engine which allows realistic driving. This is in contrast to Simulation of Urban MObility\footnote{\url{https://www.eclipse.org/sumo/}} (SUMO) which focuses on large scale movements of vehicles but with very simplified physics. In contrast our model is for a vehicle with egocentric coordinates, where the physics of the car and the road are integral parts of the model. In our simulation vehicles are affected by gravity, road and tyre friction and other forces which can be adjusted as needed. For example, when the autonomous vehicle accelerates or brakes, the specified speed is not reached immediately but takes time (depending on the vehicle's parameters), as it would in the real world. While the model presented here is still a simple proof of concept, it highlights the importance of physics. In this respect we are closer to the Open Dynamics Engine\footnote{\url{https://www.ode.org/}} (ODE) than large scale traffic simulators such as SUMO.

There are many tools and components available with Unity that can alter an object's behaviour. Modifying and combining them is easily achieved by adding scripts to the objects. Each object's functionality and logic is controlled via attached scripts written in C\#. All of the logic of the autonomous vehicle in our simulator is located in the scripts attached to the autonomous vehicle object, while the logic of other traffic participants is located in scripts attached to the individual participants.

\section{Methodology}
\label{methodology}
Our overall approach is illustrated in Figure \ref{overallApproach}.
Our Promela models and Unity simulator are described in detail in Sections \ref{sect:promelamodels} and \ref{sect:unitymodels} respectively and are available to download from our git repository \cite{Pagojus2021}.

To explain our approach we start on the left hand side of Figure \ref{overallApproach}. The simulator represents the autonomous vehicle, the road and other vehicles. If a vehicle from the left lane disappears from view, a new one is spawned in front of all vehicles in the left lane (see Section \ref{usingmodelchecking}). The autonomous vehicle senses its own position and the position of any vehicle either ahead of it in its own lane (travelling in the same direction) or in the other lane, heading towards it, within a given range (how this range is determined is discussed in Section \ref{sect:unitymodels}). This information is extracted from the simulator as a set of sensor readings and combined with a prepared template Promela specification to produce a full Promela specification representing the system. Spin is used to check the model for violations of the LTL property \texttt{!<>p}, where $p$ is a proposition stating that an overtake has occurred.  This property states that {\em in all paths an overtake never happens}. Note that the ``for all paths" part of the property is implicit for all LTL properties so is not included. The first path encountered which does not satisfy the property (i.e. for which an overtake does take place) will be returned as a trail file consisting of a path (states and transitions) leading to the property violation. We refer to this as a {\it solution path} (rather than an error path) as in our case finding a path that {\it doesn't} satisfy the property given, is a solution to our task. Note that our property is not time-limited (we can not express timed properties in LTL). This is sufficient for our purposes, but in Section \ref{sect:implementation} we discuss measures taken to ensure that the generated solution is {\em sensible}.

Spin is then used to execute the solution path (via a {\em guided simulation}). During this simulation we monitor actions taken by $AV$ corresponding to the (non-italic) labels on edges in Figure \ref{fig:finiteStateMachine}. These actions are then returned as an ordered list to the Unity simulator, so that the autonomous vehicle can now carry out a successful overtaking manoeuvre. 

\begin{center}
\includegraphics[angle=90,origin=c,width=0.7\textwidth]{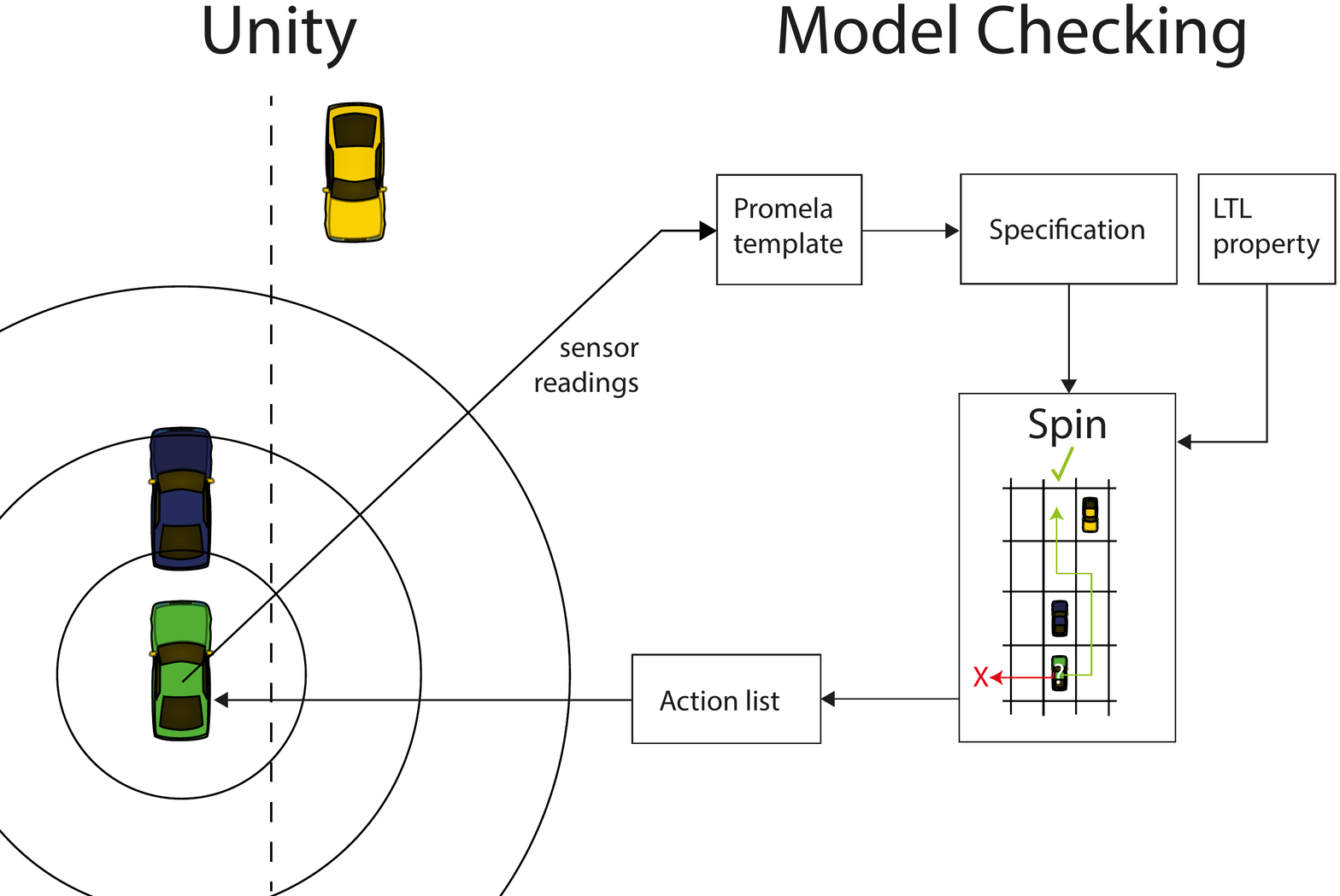}
\captionof{figure}{The Unity model provides a realistic simulation and sensor readings from the autonomous vehicle. These sensor readings are used to adapt a Promela template into a specification which is verified to find a solution path. The path is executed using Spin to extract a list of actions for successful overtaking.}
\label{overallApproach}
\end{center}

\section{Promela Models}\label{sect:promelamodels}
Our main Promela model is generated from a template, and consists of a single process representing the autonomous vehicle ($AV$). Initial values (obtained from the Unity output - see Section \ref{sect:unitymodels}) representing the current positions of any other vehicles present within a defined distance of $AV$ are set. These include any vehicles in front of $AV$ in the same lane, ($FV_{1}, FV_{2}, \ldots, FV_{r}$), where $r$ is the number of such vehicles, and a single vehicle approaching $AV$ in the other lane (i.e. an oncoming vehicle) $OV$. An example of how such an initial setup can be visualised is shown on the left hand side of  Figure \ref{finalmovementexample}. In this section we assume that $r\leq 2$. However, as discussed in Section \ref{sect:unitymodels} our simulation setup can cope with $r=3$ (limited only by the number of vehicles the sensors can detect).

\subsection{Transition system and road layout}
The behaviour of $AV$, as modelled in our Promela specification, is based on a transition system shown in Figure \ref{fig:finiteStateMachine}. The transitions are labelled with boolean conditions (in italics) and actions. The conditions have the form $var==true$ or $var==\mathit{f}alse$, for a boolean variable $var$. Note that when the boolean variables $le\mathit{f}t\_lane$, $crashed$ or $overtake\_complete$ are true, $AV$ is in the left lane, $AV$ has collided with another vehicle, or the overtaking operation has been completed respectively. A successful overtake is a sequence of actions that result in $AV$ being in the left hand lane in front of all of the vehicles in the left lane that were originally ahead of it, without having crashed. These variables have initial values \texttt{true}, \texttt{false} and \texttt{false}. Transitions that are labelled by a proposition about these variables are only enabled when the proposition evaluates to true. All other transitions are assumed to be always enabled. Transitions that are labelled by actions (resetting the value of $le\mathit{f}t\_lane$, or actions $move\_\mathit{f}orward$, $\mathit{f}ast\_\mathit{f}orward$ or $slow\_\mathit{f}orward$) result in the associated actions being executed. 

\begin{center}
\includegraphics[width=0.9\textwidth]{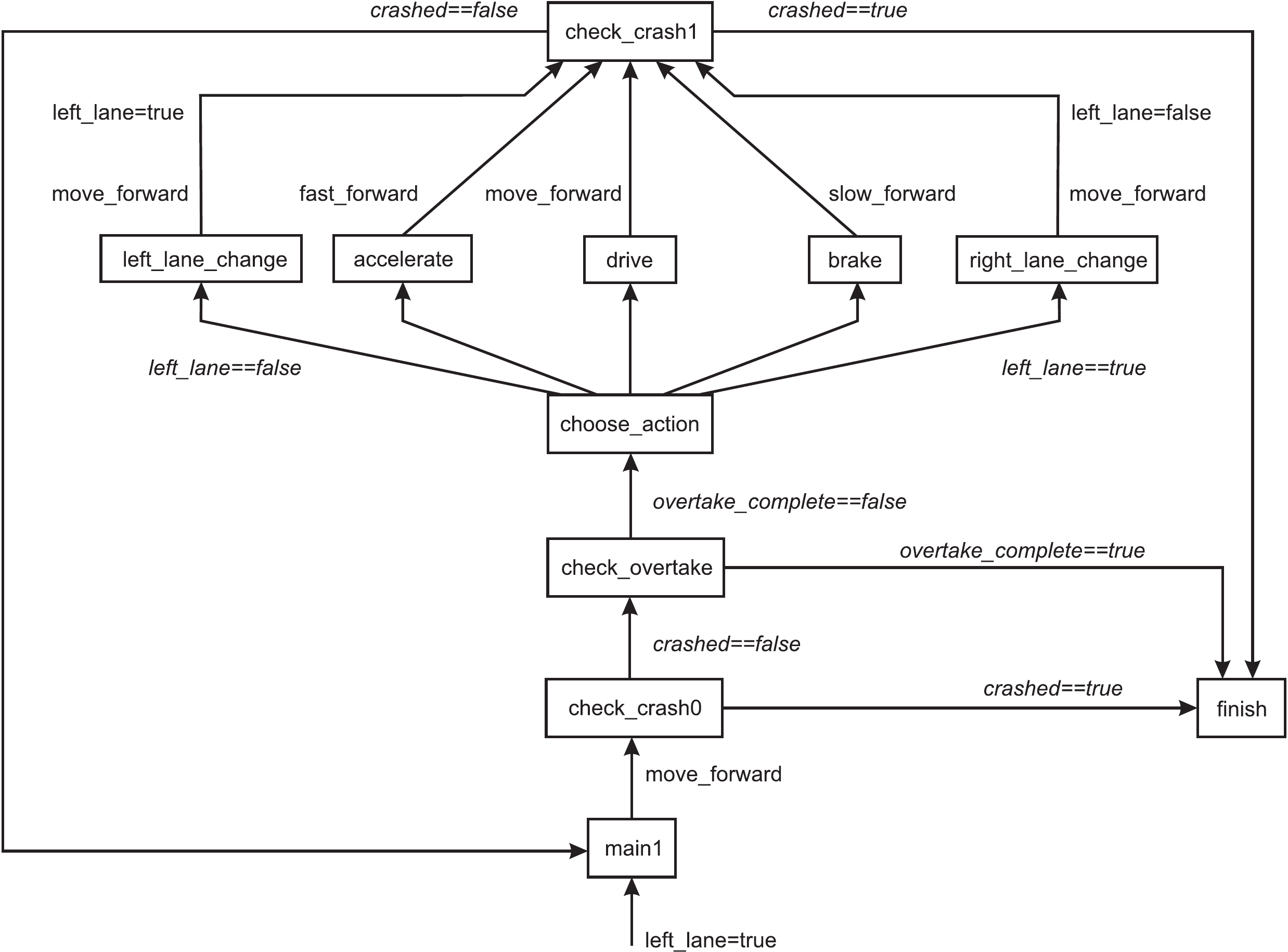}
\captionof{figure}{A transition system representing $AV$ behaviour.}
\label{fig:finiteStateMachine}
\end{center}

When $AV$ is at state \texttt{main1}, it moves forward at default speed - indicated by the $move\_\mathit{f}orward$ action). Unless a crash has occurred or a successful overtake manoeuvre has been completed it then progresses to one of five states (depending on whether any associated boolean condition is satisfied). This state is either \texttt{drive}, in which case $AV$ moves forward again at default speed; a lane-changing state, in which case $AV$ changes lane and moves forward at default speed;   \texttt{accelerate} in which case $AV$ moves forward at increased speed - indicated by the $\mathit{f}ast\_\mathit{f}orward$ action); or \texttt{brake} in which case $AV$ slows down (i.e. continues to move forward but at reduced speed) - indicated by the $slow\_\mathit{f}orward$ action. After an action has been chosen a check is made from the \texttt{check\_crash} state to determine whether $AV$ has collided with another vehicle. If a crash has occurred $AV$ moves to the $\textit{f}inish$ state and no further actions are taken. Note that the check for a completed overtake will not happen when a crash has occurred, so a successful overtake is assumed to be one without a collision.


The size of the road grid squares is chosen to represent the distance moved in one time-step when travelling at default speed. Spin is a finite state model checker, so a suitable abstraction to integer values is necessary.

\begin{figure*}
\begin{center}
\includegraphics[width=0.65\textwidth]{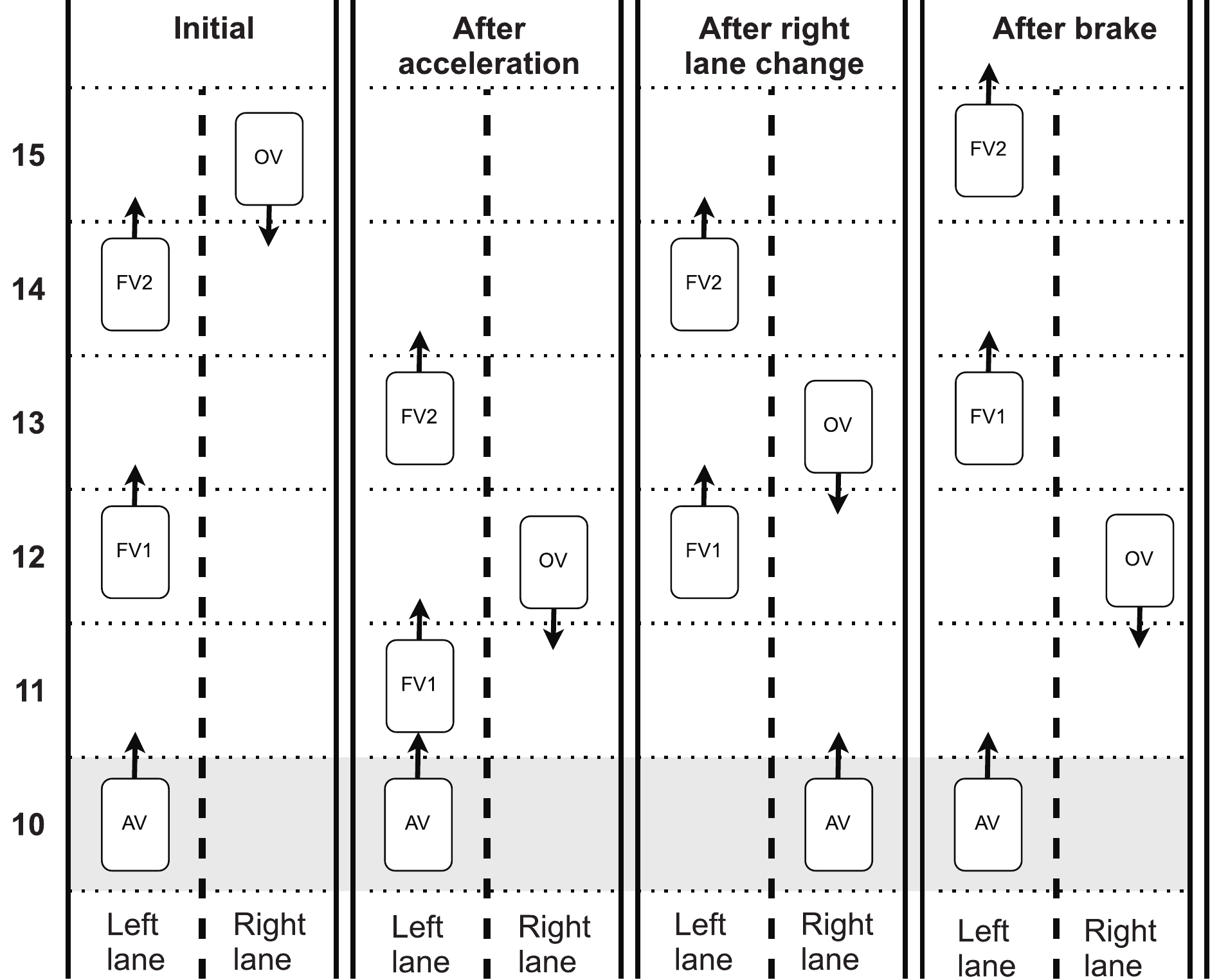}
\caption{\label{finalmovementexample} Road with example initial car positions (Left) and the resulting car positions from this initial layout after acceleration, right lane change and brake actions are executed from the initial positions respectively.}
\end{center}
\end{figure*}

All vehicles other than $AV$ always move forward one position per time-step in their direction of travel (whereas $AV$ can move a variety of positions per time-step, depending on which action it is taking). All movement takes place with respect to $AV$, whose initial vertical position is $10$. This position is chosen so that there is sufficient space behind $AV$ to place the $OV$ when it has passed and is still within range of the sensors. The result of all actions by $AV$ is seen as an effect on the other vehicles, while the vertical position of $AV$ remains fixed (although it may change horizontal position - i.e. change lane). For example, any $move\_\mathit{f}orward$ action will have no effect on  $AV$ or on any of the vehicles $FV_{i}$ as their position with respect to $AV$ is unchanged (assuming that $AV$ has not previously accelerated). However $OV$'s position will change by two positions. Centering the coordinate system on $AV$ has the additional benefit that the model becomes independent of the road curvature as long as an on-board lane estimation algorithm is available (as described in \cite{deo2018}).
 
As a shorthand, we refer to the {\it actions} of $AV$ by the names of the associated states from which the forward actions are executed, namely  \texttt{left\_lane\_change}, \texttt{right\_lane\_change}, \texttt{accelerate} and \texttt{brake}. 
 
In Figure \ref{finalmovementexample} we illustrate the effect of the \texttt{accelerate}, \texttt{right\_lane\_change} and \texttt{brake} actions on the (relative) positions of the vehicles from the example initial positions shown on the left of the diagram. For the \texttt{accelerate} action, in one time step, as $AV$ moves two positions forward and the $FV_{i}$ only one, the relative positions of the $FV_{i}$ are reduced by $1$. Similarly as $OV$ moves down one position while $AV$ moves up two, the overall effect is for $OV$ to move three positions closer to $AV$. For the \texttt{right\_lane\_change} action, $AV$ moves to the right lane and moves forward one position. As the $FV_{i}$ also move forward one position their (vertical) distance from $AV$ is unchanged. At the same time $OV$ moves one position towards $AV$ so their vertical distance from $AV$ is reduced by $2$. The \texttt{brake} action takes place over two time steps (as $AV$ moves forward at half speed). In this period $AV$ moves forward one position and the $FV_{i}$ move forward two, so their relative distance from $AV$ increases by $1$. Similarly $OV$ moves two positions towards $AV$ in the time it takes for $AV$ to move one position towards $OV$, and so the relative distance between them decreases by three.

\subsection{Implementation}\label{sect:implementation}
Our Promela model consists of a single proctype which implements the behaviour of $AV$ and updates the positions of the other vehicles accordingly. The initial grid position of $AV$ is assumed to be $10$, and the relative positions of $FV_{1}$ and, if relevant, $FV_{2}$ and $OV$ are determined from the Unity output.  How the number of grid positions are chosen, and how real values from the Unity sensor readings are mapped to these integer positions are explained in Section \ref{sect:unitymodels}. 

We include the declaration of a boolean proposition $p$ defined as $overtaken==true$. The variable $overtaken$ is originally set to false, but set to true when a successful overtake manoeuvre takes place - i.e. when $AV$ is in the left lane and the position of $FV_{1}$ (and, if relevant of $FV_{2}$) is less than $10$. Note that the value of $p$ is only updated after a check for a collision has taken place, so we can assume it is only true if no crash has occurred. In order to find a path in which a successful overtake occurs, we use Spin to disprove the LTL property \lstinline{!<>p}, i.e. {\em in every path, p is never true}. Any counter-example to this property will consist of an error path in which $p$ becomes true - i.e. a successful overtake takes place. Note that after every action a check is made to determine whether a crash has occurred. If it has, the path is abandoned (and not be returned as an error path). We know therefore that any returned path is collision-free. Spin verification will report an error indicating that a path has been found for which a successful overtake has occurred.  If no error path is found then no  overtake is possible. The error path (consisting of a list of states and transitions visited on the path) can be explored via a simulation using Spin. This is known as a {\it guided simulation} as it is guided by the latest error path. 

After any action, if the position of any of the other vehicles is the same as that of $AV$, a crash has occurred. In this case $AV$ will progress to the \texttt{finish} state, and the path is aborted. This is perfectly legitimate but will not happen in any error path returned (as $p$ is not updated until never true in any path for which this happens). 

A \texttt{do\ldots od} statement (see Section \ref{sect:promela}) is used to represent the choice between the transitions to  (states) \texttt{left\_lane\_change}, \texttt{accelerate}, \texttt{brake}, \texttt{drive}, and  \texttt{right\_lane\_change}. This choice is non-deterministic - any enabled transition (i.e. transition whose condition is true, or which has no condition) can be selected. It is the choices made via this statement that will provide the action list that is returned to the Unity simulation, so it is important that we can retrieve them. By annotating our Pomela code with suitable \texttt{printf} statements we can ensure that strings indicating the choices are contained within the guided simulation. These strings can then be extracted, in the order in which the corresponding actions are taken, during the guided simulation.

From hereon, we will refer to a {\it solution path} rather than an error path (as the error path is in fact providing a solution to our problem). A difficulty with retrieving solution paths in this way is that Spin will return the first path it encounters that violates the property, which may not be a {\it sensible} one (it can also return the shortest path if required, at a cost). For example, Spin may return a path in which $AV$ swaps lane repeatedly before executing an overtake manoeuvre, despite the fact that a safe overtake would have been perfectly possible without this behaviour. The returned path is legitimate, but a more sensible path would be preferable. Theoretically, as a \texttt{do\ldots od} statement represents non-deterministic choice, any solution path is viewed as equally legitimate. However, we can influence the solution found by choosing our ordering of statements within the \texttt{do\ldots od} in such a way that a search of the state-space will find a sensible solution path before a pathological one is explored. We have found that the best ordering is as follows: 
\texttt{left\_lane\_change}, \texttt{accelerate}, \texttt{right\_lane\_change}, \texttt{drive}, \texttt{brake}. Our sensible solutions avoid unnecessary lane changes, repeated braking and staying for an unnecessarily long time in the right lane. 

We can also restrict the behaviour of $AV$ by limiting the number of lane changes. This again helps to direct the model checker to a sensible solution. 

\subsection{Danger Zone and Preparations Model}

Our model uses a fairly coarse discrete representation of space. While this is generally adequate, it can lead to unsafe behaviour when $AV$ judges its distance from the $OV$. Whether the $OV$ is in reality at one end of a grid square or the other can mean up to three seconds difference in the amount of time $AV$ has to safely carry out an overtake manoeuvre (see Section \ref{sect:unitymodels}).   
To solve this problem, we introduced a {\em danger zone} around $OV$ - i.e. $AV$ was not permitted to enter a grid position either side of the position containing $OV$. We refer to this version of our model as \texttt{final\_model}.

After running multiple simulations of our model we discovered two unexpected situations in which $AV$ failed to find a way to safely overtake two vehicles when it should have been able to:
\begin{itemize}
    \item Case A: Having overtaken $FV_{1}$ there is no gap to return to the left lane as $FV_{2}$ is too close. There appears to not be enough time to overtake this second vehicle due to the closeness of the danger zone around $OV$. A viable solution would be to brake and return to the left lane behind $FV_{1}$.  However, although there might be enough distance between $AV$ and $OV$ for such a manoeuvre, the danger zone may prevent it. Thus although this might be the best possible course of action the braking manoeuvre would be rejected and a crash would be inevitable. 

    \item Case B: After overtaking $FV_{1}$, $AV$ may not start accelerating to get closer $FV_{2}$ while $OV$ is far away, and so will not be at an optimal position to overtake $FV_{2}$.
\end{itemize}

We were not able to integrate solutions to these two corner cases in \texttt{final\_model}. Instead we created a similar model called the \texttt{preparations\_model} with different goals and slightly changed constraints from those in the \texttt{final\_model}. When \texttt{final\_model} fails to find a path, depending on which of these two cases caused the failure, a version of \texttt{preparations\_model} is generated and used to find a path. In each case a path is now used to return $OV$ to a safe situation from which to rerun  \texttt{final\_model}.

To solve case A, we have removed the danger zone rule from the \textit{preparations} model and adjusted the goal of the model checker from finding a path to overtake $OV$, to finding a path to get back into the safe position - any gap on the left lane. Removing the danger zone introduces risks, but in a real-world emergency situation it is hard to predict the movement of an oncoming vehicle and important to do everything to avoid a head-on collision.

To solve case B, we change the goal to finding a path that closes the gap between $AV$ and the $OV$, thus preparing $AV$ for the most efficient overtaking manoeuvre.

Since there are two different goals in the \textit{preparations} model, we have to select which one to pursue. We have noticed that in case A, $AV$ is always positioned in the right hand lane, and in case B $AV$ is always positioned on the left hand lane. Hence, when \textit{preparations} model is invoked we can use this insight to select the appropriate goal.

\section{Simulator}\label{sect:unitymodels}
\begin{figure*}
\includegraphics[width=\textwidth]{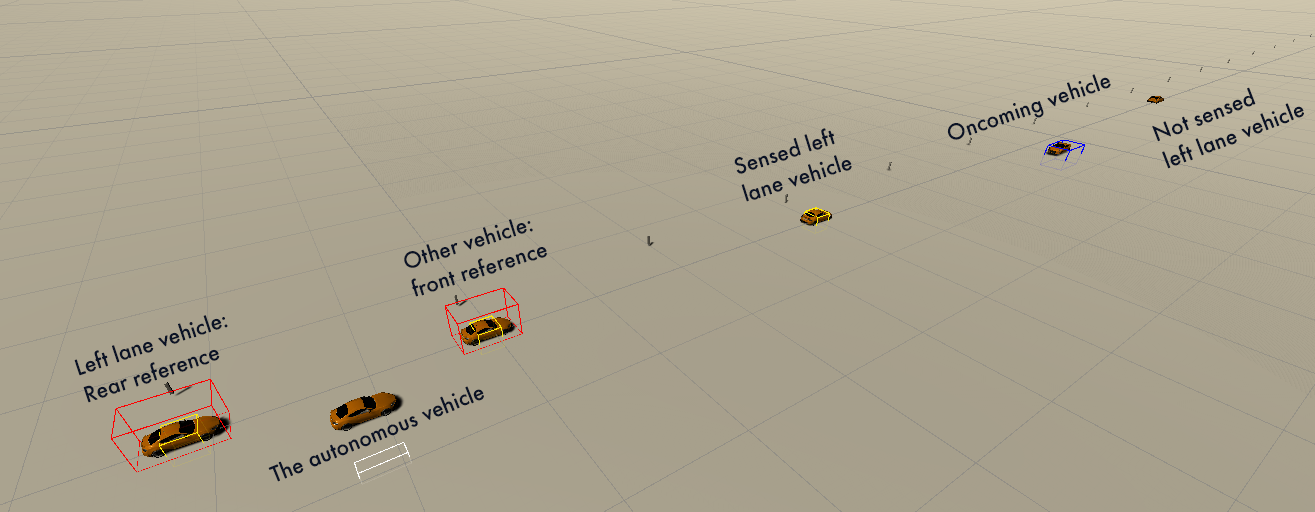}
\caption{\label{fig:singlesimulatorview} The simulator view. $AV$  is overtaking two consecutive vehicles. Vehicles with yellow boxes around them are sensed by the 360-degrees sensor. Vehicles with red boxes around them are used as reference points for the acceleration and braking manoeuvres. $OV$, marked with a blue box, is sensed using the long range sensor.}
\end{figure*}

In this section, we cover the key features of our Unity simulator: the vehicle model, space discretisation and sensors. We continue to use the terms $AV$, $OV$, etc. to identify the vehicles. A view from the simulator (in which $AV$ is overtaking $FV_{1}$ and $FV_{2}$) is shown in Figure \ref{fig:singlesimulatorview}.

The vehicle model used in the simulator was from the Unity Asset Store\footnote{\url{https://assetstore.unity.com/packages/3d/vehicles/land/low-poly-sports-car-20-144253}}. It is affected by Unity's physics engine and can be controlled by adjusting throttle, brake and steering inputs. Instead of adjusting the position of the vehicle directly, we manipulate these inputs to simulate realistic vehicle movement. Note that the states of the Unity model, though having similar names to those we use in our Promela model, are not identical (hence we use capital initial letters). 

We set the maximum speed of $AV$ to 25.2 km/h in a \texttt{Drive} state which is a speed that is often used for vehicles in this context \cite{Althoff2014, li2016}. We limit the speed of $AV$ to at most 50.4 km/h in the \texttt{Acceleration} state and to at least 7 km/h in the \texttt{Brake} state. 
We set the speed of all other vehicles to be constant at 25.2 km/h.

We divide the continuous world into 21-meter long road segments. $AV$'s manoeuvres (e.g. lane change and acceleration) take around three seconds to complete in the simulator. A vehicle moving at a speed of 25.2 km/h will cover 21 meters in three seconds. This makes it easier to match the behaviour of the simulator and the Promela model.

An autonomous vehicle senses the environment using sensors attached to it.
Initially $AV$ had 6 distance sensors: one  for each side of the vehicle; two in front - one for sensing the left lane and one for the right lane; and one for sensing left lane vehicles directly in front of $FV_{1}$.
As we introduced more traffic, such a system was insufficient to accurately sense all visible vehicles on the left lane.
We kept the right lane and right side sensors for detecting right lane vehicles as they were, but for left lane vehicles, we simulated a 360 degrees sensor. We used Unity's SphereCast\footnote{\url{https://docs.unity3d.com/ScriptReference/Physics.SphereCast.html}} to detect the positions of all left lane vehicles relative to $AV$ within a fixed 100 meter radius and then cast a laser (Unity's RayCast\footnote{\url{https://docs.unity3d.com/ScriptReference/Physics.Raycast.html}}) to each of those to determine which vehicles were in a direct line of sight. The result can be seen in Figure \ref{fig:singlesimulatorview}. Existing LiDAR sensors like Velodyne Alpha Prime (VLS-128)\footnote{\url{https://www.mapix.com/lidar-scanner-sensors/velodyne/alpha-prime/}} or Velodyne  HDL-64E\footnote{\url{https://www.mapix.com/lidar-scanner-sensors/velodyne/velodyne-hdl-32e/}} used in \cite{li2016} are capable of sensing a high resolution 360 degree view of up to 245m and 120m respectively which is essentially what we have achieved using Unity's provided functions. 

The list of vehicles in sight is used by $AV$ to find the positions of the closest vehicle behind and in front of it. These are used as reference points for its movement (for example to know when it has moved one position closer to $FV_{1}$ when accelerating).
We set the right lane sensor to have a range of up to 357 meters (17 road segments in front of $AV$). Having such a long range allows $AV$ to sense $OV$ early enough to be able to overtake three consecutive vehicles. Overtaking three vehicles is not a common manoeuvre and having 357 meters of a straight road that can be sensed from the on-board sensor is only usually possible on a motorway. In theory, however, it should be possible to do this using sensors such as Acuity AR2000\footnote{\url{https://www.acuitylaser.com/product/laser-sensors/long-range-sensors/ar2000-laser-distance-meter-ar2000/}}. Sensing so far in front of $AV$ has helped us to test the capabilities of the model checking algorithm as multiple vehicles had to be considered at once. For overtaking a single vehicle, $OV$ has to be sensed when it is at least 10 segments (210 meters) away as discussed above. In that case, in theory, all the sensing could be done using a previously mentioned Velodyne Alpha Prime (VLS-128) LiDAR sensor. 

Our final sensor system is as follows:
\begin{itemize}
    \item 360 degree sensor with 100 meter range for sensing left lane vehicles,
    \item Front Right Lane Sensor with 357 meter range for detecting the Oncoming Vehicle,
    \item Right side sensor for sensing whether there is an obstacle on the right side of the Autonomous Vehicle that the Front Right Lane Sensors could not capture.
\end{itemize}

\section{Combining the Unity game engine and model checker}
\label{usingmodelchecking}
The verification of our model takes about 20 milliseconds. However, the Promela model has to be compiled to  C code before it can be verified, which takes around 3 seconds. Because of this additional time requirement, it is currently not possible to use the Spin model checker based algorithm in real-time. For that reason, we pause the simulation while the solution path is generated, in the hope that faster compilation times can be achieved in the future.

While, ideally, verification should be done as often as possible for the safest and up to date trajectory, it causes a few problems.
Recomputing the trajectory too often requires the simulation to be paused repeatedly, resulting in slow simulation. It can also make $AV$ susceptible to noise and to generating over-reactive actions \cite{li2016}.  Recomputing the trajectory in the middle of an overtaking manoeuvre would make $AV$ abandon the manoeuvre (by emergency braking) if $OV$ appeared closer than it had been before the manoeuvre had started - even if the manoeuvre could have been completed successfully. 
To solve these problems, we only ran the verification when either new surrounding vehicles were sensed or when no generated actions remained. 

We refer again to Figure \ref{overallApproach} to explain how the whole process of creating the model, verification and returning the action list works. First
$AV$'s sensory information is discretised, and the  \texttt{final\_model} created from a Promela template. This is then passed to Spin to compile and verify. 
If a solution path is generated, it is executed using Spin's guided simulation facility. The actions of $AV$ are returned in the order they were visited in the path to the Unity simulator and  $AV$ then executes these actions in Unity. 
If no solution path is returned, the same procedure is performed with the \texttt{preparations\_model}. If a path is still not returned then we conclude that there is no path that will result in successful overtake without a collision. The simulation is halted and a failure noted. This situation is extremely rare, and did not occur during our experiments for which results are presented in Section \ref{sect:results}.


To test our system we created a Unity scene where traffic is automatically and randomly generated. The scene contains a constant number of vehicles - once a vehicle falls behind $AV$ by 5 positions it is removed and a new one is spawned.

If a vehicle from the left lane is removed, a new one is spawned in front of all vehicles already in the left lane. There is an equal chance of the vehicle being spawned one, two, three or four positions in front of the front-most left lane vehicle.  An additional constraint prevents these vehicles from forming a sequence of more than three consecutive vehicles without an empty position in between them, as this would prevent $AV$ from being able to overtake them all. 

Once removed, right lane vehicles are spawned behind all remaining right lane vehicles. The position of the new vehicle is either $8, 12, 16$ or $20$ positions behind the last vehicle (with probability $1/8$, $1/4$, $1/4$  and $3/8$ respectively).  We used bigger gaps in the right lane to provide enough space for overtaking.

Once a simulation is started, $AV$  continues to overtake left lane vehicles using the action lists generated by the model checker. The simulation is stopped when $AV$ hits another vehicle or when the model checker fails to produce a path. Each time the simulation is stopped, we note the simulation time, the number of overtaken vehicles, the distance travelled and the cause of the failure.

\subsection{Results}\label{sect:results}
\begin{table*}
\begin{tabular}{|m{1,5cm}|m{1,5cm}|m{1,5cm}|m{1,5cm}|l|} \hline
\emph{Simulation ID} & \emph{Runtime} & \emph{Vehicles overtaken} & \emph{Distance Travelled (km)} & \emph{Failure cause} \\ \hline
1 & 5h 30m & 480 & 92.8 & Sensor failure (two vehicles at position 9)
 \\ \hline
 
2 & 2h 20m & 230 & 43.0 & Sensor failure (fake gap)  \\ \hline

3 & 2h 00m & 200 & 38.2 & Sensor failure (2 vehicles at position 11)  \\ \hline

4 & 5h 20m & 515 & 102.3 & Sensor failure (fake gap)
 \\ \hline
 
5 & 1h 00m & 100 & 19.6 & Sensor failure (fake gap)
 \\ \hline
 
6 & 2h 30m & 260 & 52.5 & Steering failure
\\ \hline

7 & 1h 30m & 149 & 28.7 & Sensor failure (rear sensor sees an extra obstacle)
\\ \hline

8 & 3h 10m & 296 & 60.5 & Sensor failure (fake gap)
\\ \hline

9 & 0h 40m & 64 & 12.0 & Sensor failure (fake gap)
\\ \hline

10 & 2h 30m & 105 & 21.9 & Sensor failure (fake gap)
\\ \hline

11 & 4h 30m & 229 & 43.8 & Sensor failure (rear sensor sees an extra obstacle)
\\ \hline

12 & 0h 30m & 44 & 10.0 & Sensor failure (rear sensor sees an extra obstacle)
\\ \hline
 
\end{tabular}
\caption{\label{tab-res}Simulation results.}
\end{table*}

We ran the simulation twelve times until it stopped, meaning that there were twelve failures in total. Note that all of these failures were observed in the simulation - they were not failures to produce a safe overtaking manoeuvre by the model checker. A proposed manoeuvre was always generated. Simulation time does not include the time during which the simulator was paused for the model checking to take place (the clock was stopped during this time). The results of individual runs can be seen in Table \ref{tab-res}. The total simulation time was 31h 20m in which $AV$ overtook 2672 vehicles and covered 525.34 km. This means that, on average, $AV$ performed five overtakes per kilometer. The median distance covered without a failure was 40.6 km. The median number of vehicles overtaken before crashing was 214.5. The three top reasons for a crash were: sensing an empty road segment between two consecutive left lane vehicles when there was none (six instances); sensing a left lane vehicle when there was none and thus failing to find a free gap to get back into the left lane to let the $OV$ pass (three instances); and incorrectly sensing two left lane vehicles at the same road segment  (two instances). There was also one instance where there was a steering failure causing $AV$ to spin. Note that sensor failures in the simulator are due to the way that Unity implements its laser range finder.

In order to make our experiments faster, $AV$ was forced to do as many overtakes as possible and in 40.6 km it did 203 overtakes on average. This is much more than an average vehicle would do in reality. Having fewer overtakes per kilometer would yield much greater distances before failure.

\subsection{Discussion}\label{sect:discussion}
We have shown that it is possible to use our model checking technique 
for planning an overtaking manoeuvre. This process can viewed as a reward-based task which could be solved with reinforcement
learning \cite{Sutton98}, most prominently deep reinforcement
learning \cite{Mnih2015} and deep policy gradient
learning \cite{Yu2020}. However, as with any (deep)
neural network approach the error rate will decline but will never be
zero - even with perfect sensor data. This is a major problem of
neural network based systems in mission critical applications where a
hard decision with perfect sensor data should lead to completely reliable
outcomes \cite{Kroll2017}. Generating a plan using model checking should guarantee that the plan will succeed. 

Deep neural network approaches have the advantage that they offer end-to-end learning from pixels to actions \cite{Mnih2015}.
However, while this approach is attractive,  it has also been criticised as
pure correlation-based learning lacking a deeper understanding that
symbolic descriptions could potentially offer \cite{Marcus2020}. While
our approach is very simple, it demonstrates, as a proof of concept, how
analogue information such as speed, timed sensor events
and actions can be converted into a symbolic representation, in our case the Promela
model specification language. 

While contemporary deep neural reinforcement algorithms can operate in a continuous
action space \cite{Lillicrap2016}, finite state model checking relies on a discrete action space.
Our division of the road into 21 meter long segments,
though sufficient to represent rural roads or empty motorways where vehicles maintain long distances between each other, would not be appropriate for congested locations like cities. For comparison, in related work on the assessment of autonomous cars \cite{Althoff2011, Althoff2009} the road is divided into 5 meter segments. Using a finer discretisation would not only allow us to represent the world more accurately, but it would allow us to model vehicle speed more precisely. In \cite{Althoff2011} prediction results 
for two Markov Chain models with different levels of discretisation were compared. The model with the finer resolution performed much better but had a much longer computation time.
Using finer resolution would require Spin to explore a much bigger state-space which would make the real-time application too slow. 

Sensors are not one hundred percent reliable \cite{aldsugyon2016} and our analysis shows that crashes were most often caused by inaccuracies in the sensor measurements. Instead of trying to make sensors in the simulation completely accurate, any improvement should include the ability to detect and correct sensor failings. 

In the real world $AV$ will have to deal with highly unpredictable traffic. To reflect these uncertainties we could use a probabilistic model checker, such as PRISM \cite{kwiatkowska}. However we chose Spin for its simplicity as a first approach. An alternative approach to dealing with uncertainty would be to integrate techniques for traffic prediction like those described in \cite{Althoff2009} into our model.

The biggest problem so far is the time that Spin takes to create the C code verifier from the Promela code. In our experiments we paused the simulation while this was happening, but this was not realistic. This three second delay is  too long for a real-time application.  In \cite{li2016} it is argued that even 100 ms lag is unacceptable. We are using an off-the-shelf tool which is not specifically suited for such an application. However, we believe that in the future our approach can be applied using more efficient model checkers for which this compilation step is not necessary.

\section{Conclusion}
\label{conclusion}
We have demonstrated for the first time how model checking could be used to plan overtaking manoeuvres for an autonomous vehicle $AV$. Promela models which reflecting the behaviour of all sensed vehicles on a road when $AV$ executes a range of actions are described in detail. A Unity simulation of $AV$ equipped with long range and a 360 degree sensors able to generate sensory data mapped to discrete positions at which left and right lane vehicles are positioned relative to $AV$ is presented. We have explained how this simulated sensor data is used to create a Promela model which, when verified using Spin can generate an action plan for the simulated $AV$ to execute a successful overtake manoeuvre. Testing  has shown that our proposed approach is able to generate consecutive overtake manoeuvres, but is prone to failures when sensory data is incorrect. 

This work has been successful as a proof of concept. Despite the limitations that we have identified, the approach is promising and we will continue to develop it in future work. We intend to investigate the use of a different model checker which does not have such a compilation time cost. Possibilities include using a direct model checker such as Java PathFinder \cite{visser2003} or to implement our own stripped down model checking algorithm. We also aim to address the problems discussed in Section \ref{sect:discussion}. This includes giving more choices for $AV$, increasing resolution and vehicle speed, decreasing the compilation time and including uncertainty in the movements of other vehicles. We hope that solving these problems would make model checking a useful technique in planning a safe overtake manoeuvre trajectory in an unpredictable real-world environment.

\section*{Acknowledgements}
This work was supported by a grant from the UKRI Strategic Priorities Fund to the UKRI Research Node on Trustworthy Autonomous Systems Governance and Regulation (EP/V026607/1, 2020-2024).
Ivaylo Valkov was supported by the EPSRC Science of Sensor Systems Software grant (EP/N007565/1, 2016-2022).

\bibliographystyle{eptcs}
\bibliography{bibliography}


\end{document}